%% file: master.tex
\documentclass[letterpaper]{article}
\usepackage{aaai}
\usepackage{graphicx}
\usepackage{times}
\usepackage{helvet}
\usepackage{courier}
\usepackage{amsmath} 
\usepackage{amssymb}  
\usepackage{amsthm} 
\usepackage{alltt}
\usepackage{xspace}
\usepackage{url}
\usepackage{html}
\usepackage[table]{xcolor}
\usepackage{epstopdf}

\nocopyright

\begin{document}

\title{Strategic Planning for Network Data Analysis}


\author{Kartik Talamadupula$^\dag$\thanks{This work was carried out while the author was an intern at IBM's T. J. Watson Research Center.} \and Octavian Udrea$^\S$ \and
  Anton Riabov$^\S$ \and Anand Ranganathan$^\S$  \AND
  $^\dag$Dept. of Computer Science and Engg.\\
  Arizona State University\\
  Tempe AZ 85287\\
  \begin{normalsize}{\tt krt} @ {\tt asu.edu}\end{normalsize} \And
  $^\S$IBM T. J. Watson Research Center\\
  17 Skyline Dr\\
  Hawthorne NY 10532\\
   \begin{normalsize}{\tt \{oudrea,ariabov,arangana\}} @ {\tt
       us.ibm.com}\end{normalsize}}

\maketitle










\begin{abstract}


  As network traffic monitoring software for cybersecurity, malware
  detection, and other critical tasks becomes increasingly automated,
  the rate of alerts and supporting data gathered, as well as the
  complexity of the underlying model, regularly exceed human
  processing capabilities. Many of these applications require complex
  models and constituent rules in order to come up with decisions that
  influence the operation of entire systems. In this paper, we
  motivate the novel {\em strategic planning} problem -- one of
  gathering data from the world and applying the underlying model of
  the domain in order to come up with decisions that will monitor the
  system in an automated manner. We describe our use of automated
  planning methods to this problem, including the technique that we
  used to solve it in a manner that would scale to the demands of a
  real-time, real world scenario. We then present a PDDL model of one
  such application scenario related to network administration and
  monitoring, followed by a description of a novel integrated system
  that was built to accept generated plans and to continue the
  execution process. Finally, we present evaluations of two different
  automated planners and their different capabilities with our
  integrated system, both on a six-month window of network data, and
  using a simulator.

\end{abstract}
\vspace{-5mm}


\label{sec:intro}
\section{Introduction}




As the reach of the internet grows, network traffic monitoring
software for malware detection, cybersecurity, and other critical
applications is becoming increasingly important and is being used in
larger deployments than ever before.  Often, this increase in use
brings with it a manifold jump in the amount of data that is collected
and subsequently generated by the application, and an increase in the
importance of the processes that are used to analyze that data.
Usually, these processes are codified as sets of rules that are
applied to the incoming data. These rules are often only known to
domain experts, whose job it is to sift through the data and arrive at
decisions that will determine the future of the system as a whole.
However, given that these experts are human, an explosion in the size
of the incoming data as well as its complexity renders this manual
approach grossly infeasible and worse, susceptible to errors and
catastrophic failure. Such eventualities motivate the general task of
automating the processes that the experts undertake in analyzing the
data. We call this the {\em strategic planning} problem.

The aim of strategic planning is to consider all the components and
data at the system's disposal and come up with a sequence of decisions
(actions) to achieve, preserve or maximize some specific objective.
The strategic planning process may be closed loop, i.e., decisions are
made based on the data currently available, but may change given new
data that is sensed from the world, or otherwise provided. We look to
apply automated planning methods, and planners, as mediators to this
strategic planning problem. The advantage of framing this as a
planning problem is that we can use multiple levels of planners, and
different models, in order to represent the different facets of the
system. This enables us to expose the system to various experts and
their knowledge at different levels of detail. A single system that
takes just one model of the domain into consideration would either
fail to scale to the large amounts of data that the system must
process, or make very little sense representationally to the experts
whom the system is designed to assist.

A few previous approaches have considered employing automated planning
methods and terminology to the processing of streaming data and the
composition of system components, as a limited version of the
strategic planning problem. The closest such work is that of Riabov
and Liu~\cite{riabov2006scalable}, who developed a specialized version
of a Planning Domain Definition Language (PDDL) based approach to
tackle a real streaming problem. However, that approach focused on
producing a single analysis flow for a given goal, and did not support
investigations that involve multiple decisions based on sensing
actions. Other methods have been proposed, consisting of more
specialized approaches where composition patterns are
specified~\cite{ranganathan2009mashup}, or specialized planners like
the MARIO system are used~\cite{bouillet2009mario}, as a lower level
analog to the higher level strategic planning problem. In this work,
we seek to create an integrated system that marries lower-level
planners like MARIO -- which deal with one step of analysis at a time
-- with automated planners to conduct multi-step investigations that
assist administrators and other stakeholders. In particular, we
consider the problem of administering a large network infrastructure
with an eye towards security. Networks are typically monitored by a
set of {\em network administrators} - humans who have experience with
that network's setup and can recognize the various types of traffic
coming in and going out of the topology. These admins often have a
highly complex model of the data flow profiles in the network at any
given time. They call upon this experience in order to identify and
isolate anomalies and extreme behavior which may, for example, be
indicative of malware infections or other undesirable behavior. The
process of analyzing these anomalies is christened an `investigation',
and involves a complex series of actions or decisions that are
inter-dependent. These actions together form the {\em model} of the
domain that resides within the expert, and that is applied to the
incoming data. There are also a number of intermediate results from
the world in the form of new, streaming data that must be sensed and
accommodated into the on-going investigations. Finally, there are
cost, duration and quality tradeoffs to the various decisions that the
admins make. The aim of the strategic planning problem is to take in a
model of this application as seen by the experts, and to produce
decisions that will help the admins take actions that ensure the
achievement of the scenario goals.

The main contributions of our work are constructing, and then solving,
this strategic planning problem. First, we describe the various
characteristics of the strategic planning problem, and then present a
solution that enables the generation of faster decisions and
decision-sequences. After that, we describe the model of the domain
that we came up with, and its various characteristics and
properties. Finally, we present the novel integrated system that we
created to situate the plans produced using the model, and to solve
the strategic planning problem end-to-end. We then evaluate our system
with different automated planners, both in a real-world setting using
six months of network data captured from a medium size network, as
well as using a novel simulator we created to test the scalability of
the underlying planners.
\vspace{-2mm}



\section{Planning Methods for Strategic Planning}
\label{sec:2-solution}





The field of automated planning has seen a great deal of progress in
the past decade, with advanced search heuristics contributing to
faster and more efficient plan generation techniques. Concurrently,
the complexity of features that can be supported representationally by
planning methods has also increased. Where previously planners could
only handle simple classical planning problems, they now offer support
for time, cost, and uncertainty, to name a few important
factors. These issues are important to us, since our model of the
application domain (Section~\ref{sec:3-domainmodeling}) and
evaluations (Section~\ref{sec:5-evaluations}) consider various problem
scenarios that include all these features. The strategic planning
problem that we are interested in solving is deterministic in many
ways - the duration and cost of analytic applications that evaluate
the data can often be estimated with fairly good precision. However,
the analytics must run on real data from the world; this is the root
cause of uncertainty in the strategic planning problem. Depending on
the state of the world, and the data collected, the analytics may
return one of many possible results. These results then have an impact
on further planning. For example, in the network security scenario
that we applied our work to, analytics influence the course of an
investigation by labeling sets of hosts with certain properties. These
properties cannot be ascertained until runtime, when the data from the
world is gathered and given to the decision model that is being used
in the strategic planning process. Since the decisions that are chosen
at a given point can influence planning in the future, it is
impossible to construct an entire plan beforehand.


Automated planning has had to grapple with this problem in the past,
and solutions have been proposed to deal with the uncertain outcomes
of actions. Incremental contingency planning
methods~\cite{dearden2003incremental} solve this problem by building
branching plans. In our application domain, however, the sensing
actions can introduce new objects into the world model, which makes
planning for all contingencies in advance difficult, if not
impossible. More general approaches have been developed for modeling
uncertain action outcomes, such as
POMDPs~\cite{kaelbling1998planning}.  We found that these approaches
are not directly applicable in our setting, mainly due to the
difficulties in maintaining reasonably accurate
distributions. Instead, we needed a system that would eventually
analyze all anomalies, however insignificant, as long as they have
been flagged by the sensing actions. 

In our case, one of the more prominent features of the strategic
planning problem proves to be a saving grace: strategic planning
proceeds in repeating {\em rounds}, viz. distinct stages composed of
the following steps until the problem goals are met: \vspace{-0.5mm}

\begin{enumerate}
\item Data is gathered from the world via streams. \vspace{-1mm}
\item The best decision to make based on that data is chosen. \vspace{-1mm}
\item Analytics are deployed on that data based on the decision
  chosen. \vspace{-1mm}
\item The results of the analytics change the world and produce new
  data; back to step 1.
\end{enumerate}

\noindent For a given round, a selected fragment of the plan is used
to determine the next action given: (1) the data that is currently
available; and (2) the overall problems goals (which typically do not
change between rounds).  Once the decisions from that plan fragment
are available, we can treat the problem as a deterministic problem
with costs and durations, since the only source of uncertainty is the
results of the analytics that will be applied; and those analytics
only contribute data to the next round of planning. In some sense,
this can be seen as a determinization of uncertainty from the world
that is afforded by replanning. The analytics that are deployed are
really sensing actions, in that they initiate sensing in the world to
bring in new data (world facts). By replanning at the beginning of
each new round, the uncertainty that comes in with the data from the
world can be dealt with, and the best plan based on all the
information currently known can be generated. This plan is only
executed up until the next sensing action (analytic), upon whose
execution the world changes and replanning is employed again. This
method takes inspiration from previous successful approaches such as
FF Replan~\cite{yoon2007ff} and Sapa
Replan~\cite{talamadupula2010tist}, which show that a combination of
deterministic planners and replanning can be used as an efficient
stand-in for contingency methods. An important part of such a
technique, however, is the domain model used -- a subject that we
explore in the next section.
\vspace{-2mm}



\section{Domain Modeling}
\label{sec:3-domainmodeling}





Hitherto, the large scale automation of data-processing systems has
been handled by the creation of large inference rule-bases. These
rules are usually put in place using information obtained from domain
experts, and are applied to data gathered from the world in order to
come up with inferences. However, such rules do not accommodate the
concept of optimization when it comes to choosing which ones to apply.
Given a specific state of the world, {\em all} of the rules that can
possibly be applied will fire, regardless of whether or not that
application is beneficial to the problem objectives and metrics.
Inference rules are also cumbersome to specify, and hard to maintain.
Due to the nature of the inference procedure, each new rule that is
added to the base must first be checked for consistency; this means
evaluating it against all the rules that are currently part of the
rule-base, and ensuring that the introduction of this new rule does
not produce a contradiction.  Instead, a more declarative
representation that can capture expert knowledge about the domain in
question is a pressing need.


For the reasons specified above, we decided to use automated planning
methods and the PDDL representation used by that community while
modeling our network administration application. The former allows the
optimization of decisions based on the current world state and
specified metrics, while the latter is a declarative
representation. Our main application was a network monitoring scenario
where network data is being monitored for abnormal activity by
administrators. The structure of the network being monitored is
assumed to be finite and completely known to the admins, and various
combinations of analytical tools (known as analytics) may be deployed
by these admins to measure different network parameters at any given
time. For ease of monitoring and classification, the network structure
is broken down into constituent {\em hostsets}, where each hostset
consists of one or more hosts. At an elementary level, a hostset may
contain just a single {\em host}, which is a machine on that network that
has an
IP address associated with it. The process of identifying anomalies
proceeds in the following steps:


\begin{enumerate}
\item Apply an analytic application (a combination of analytics) to a
  hostset \vspace{-1mm}
\item Break that hostset down into subsets to be analyzed more in-depth, based on the
  result of the applied analytic \vspace{-1mm}
\item Report a hostset if it sufficient analysis steps indicate the
  attention of a network admin is necessary \vspace{-1mm}
\item Repeat until there are no hosts / hostsets left 
\end{enumerate}


\noindent The decision pertaining to which analytics should be applied to which
hostset is usually left to the network administrator, and their model
of the network as well as prior experience in dealing with
anomalies. It is this decision-making process that we seek to capture
as part of our declarative model of this domain. 

\subsection{PDDL Modeling}

After consulting with domain experts, we reached a consensus on the
format of these decisions: each decision consists of a set of
conditions that must hold (in the data) for that decision to be
applied, and a set of changes that are applied to the data upon the
application of that decision. This is very close to the
STRIPS~\cite{fikes1972strips} and PDDL~\cite{mcdermott1998planning}
model of an `action'. The first task in this pursuit was to elicit -
from domain experts - the list of objects in this application, as well
as the boolean predicates that model relationships between these
objects. In our application, we had 8 different types of objects: some
examples are {\em hostset}, which has been introduced previously as a
collection of one or more {\em hosts};\footnote{We do not consider the
  hosts themselves as individual entities, in the interests of
  scalability and efficient planning.} {\em protocol}, which indicates
the network protocol in use (e.g. HTTP); and {\em distancefunction},
which denotes the kind of function used to measure the similarity
between different hostsets or the distance between some behavior of a
host and the aggregated model of behavior across the network.

Modeling the predicates was a more time-consuming task, since we had
to identify the exact relationships between the objects, and how these
relationships connected the objects in the world to the decisions
(actions) that needed to be taken when looking for anomalies. A large
number of the predicates in the final version of our PDDL model were
unary predicates that indicated whether a certain characteristic
associated with a hostset was true in the current state: for example,
{\em (extracted-blacklist ?$s$ - hostset)} indicates whether we know
the set of blacklisted (malware) domains the hosts in the set $s$ have
contacted. In addition to these unary predicates about hostsets, we
had some $n$-ary predicates ($n > 1$) that related a hostset to other
objects in the model, like distance functions and protocols. We also
had some ` meta' predicates, which were used for book-keeping
purposes. An example of this is {\em (obtained-from ?$s_1$ ?$s_2$ -
  hostset)}, which indicates that hostset $s_1$ is refined from the
larger hostset $s_2$ directly, in one step through the application of
some analytic process. Finally, we had some $0$-ary predicates, i.e.,
predicates with no parameters, to denote things that were true
globally -- e.g. {\em (checked-global-frequent-hosts)}.

The list of decisions (or actions) was arrived at by having domain
experts run through some usecases, and identifying the decisions that
they took to label a hostset as exhibiting anomalous behavior that
needed to be escalated up to an admin. Once we had a list of such
decisions, we tried to determine the decisions that are required to go
from the standard initial state -- a giant hostset that contains all
the hosts in the network -- to one where a specific (and very small)
descendant of the initial hostset is reported as anomalous. This
enabled us to construct a causal graph~\cite{helmert2004planning}
$\Gamma$ of the domain theory as encapsulated by the network admins.
In $\Gamma$, the nodes stand for the decisions (actions) that the
admins could choose to perform -- remember that each decision
corresponds to an analytic process or flow -- and the edges denote a
causal relationship between two actions. More formally, if actions
$a_1$ and $a_2$ are related such that $a_1$ contains an add effect $e$
which is present in the condition list of $a_2$, then $\Gamma$ will
contain a directed edge from $a_1 \rightarrow_e a_2$ between those two
nodes respectively.

Subsequent to the construction of $\Gamma$, a clearer picture of the
domain dynamics emerged. We found that there were some initial `setup'
actions that every hostset had to be subject to, in order to gather,
filter and aggregate data for the analytics that would follow.  There
were 8 such actions in total, and the overall task of these actions
was to group the hostsets and subnetworks by protocol. The last action
in this sequence was a sensing action -- {\em
  sense-gather-final-protocols} -- that sensed the data from the
network streams and split the initial (single) hostset into multiple
hostsets, depending on the protocol that the individual hosts in the
initial hostset exhibited anomalies on. In our domain, we considered
three protocols that network data was being transmitted and received
over: {\em HTTP}, generic {\em TCP}, and {\em SMTP}\footnote{Although
  there are obviously many more protocols to consider, we found that a
  large majority of the behavior domain experts were looking to detect
  shows traces over either HTTP or SMTP, or can be found going below
  that to the TCP layer.}. Thus, after the execution of {\em
  sense-gather-final-protocols}, the graph $\Gamma$ splits into three
different branches: one each for investigating anomalies along the
three different protocols mentioned above. On each one of these
protocol-dependent paths, there are more regular actions which process
the data contained in the hostsets. There also further sensing
actions, which are decisions whose outcomes are uncertain and resolved
in the world via the procedure described in
Section~\ref{sec:2-solution}. The three different branches merge into
the last action of the causal graph $\Gamma$, viz. {\em pop-to-admin}:
this action pops-up the refined hostset at the end of the three
branches to the network administrator for further review and action,
thus fulfilling the planner's role of offering suggestions to the
human experts while cutting down on the size and complexity that they
have to deal with.

\begin{figure}[htp]
  \centering
  \includegraphics[keepaspectratio, width=0.47\textwidth, clip]
  {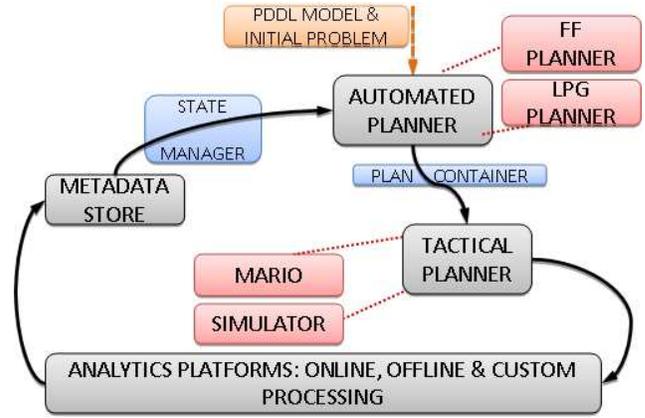}
  \vspace{-0.25in}
  \caption{The integrated system that solves the strategic planning problem.}
  \label{integrated-system}
\end{figure}

\noindent \textbf{Advanced Features}: Once the causal structure of the
domain model was established, we turned our attention to other domain
characteristics that are important to a real world application: time,
and cost. Modeling time is essential because many of the analytics
that form the backbone of the decisions suggested by domain experts
and/or the planner take time to execute on real data, and this needs
to be taken into account when generating sequences of these decisions.
Fortunately, PDDL offers support for temporal
annotations~\cite{fox2003pddl2}, and we were able to extract estimates
on action execution time from domain experts. Minimizing (or in some
way optimizing) the makespan of plans for dealing with network
anomalies is an important problem in industry, because it enables
network admins to more quickly focus their attention on pressing
problems that may infiltrate or take down an entire network
infrastructure. A similar situation exists with cost -- different
analytic processes incur different computational costs when run, and
it is useful to find a sequence of decisions / analytics to run that
costs the least while still aiding the network admins in detecting
anomalies. Since PDDL supports metric planning with costs as well, we
obtained cost estimates for each of the actions from the network
admins and annotated the actions with these. We then ran the data and
problem instances thus generated with a minimization on the
`total-cost' function as defined by PDDL.  \vspace{-2mm}



\section{Integrated System}






The domain model described in the previous section can be used with
any automated planner that supports the PDDL representation in order
to devise plans that can identify anomalies in given sets of
hosts. However, that automated planner must be part of a larger
integrated system that can process data from the world, and that can
effect the decisions that the planner suggests back in that dynamic
world. The object of the novel integrated system that we designed
around an automated planner was two-fold: (1) to translate the
decisions that are generated by the planner after taking the scenario
goals and metrics into account into analytics that can be run on
streaming data from the world; and (2) to translate the results of
analysis in the world back into a format that can be understood by the
planner. The purpose of these two objectives is to enable the
strategic planning {\em rounds}, defined in
Section~\ref{sec:2-solution}.

The schematic of our integrated system is presented in
Figure~\ref{integrated-system}. The PDDL model described in
Section~\ref{sec:3-domainmodeling}, along with the initial description
of the specific problem (hostset) under investigation, are fed to the
automated planner. This results in a plan being generated and passed
through the {\em plan container}. The decisions that were generated as
part of that plan are then handed to the {\em tactical planner}, which
decides which analytics to use to implement those decisions. The
analytic processes are then scheduled and executed on underlying
middleware platforms. The results of this execution -- in the form of
changes to the world -- are passed back on to the metadata store,
which is a module that stores all of the data related to the problem
instance. This data is then piped through the state manager, which
translates it to a representation that can be processed by the
automated planner module. In this manner, the strategic planning loop
is set up to support the various rounds. In the following, we describe
each component of the integrated system - along with the role it
performs - in more detail. \vspace{-2mm}
 

\subsection{Automated Planner}

The automated planner is the central component of our integrated
system, since it enables the strategic planning process by generating
intelligent, automated decisions based on data from the world. As
discussed in Section~\ref{sec:3-domainmodeling}, the automated
planner must support certain advanced features in order to deal with a
real world application scenario like network administration. Since
this scenario had both temporal and metric issues associated with it,
we decided to use two planners that are well known for handling these
issues. We created two distinct instances of the domain model from
Section~\ref{sec:3-domainmodeling}; one that took cost issues into
account, and another that considered temporal factors such as the time
taken to execute each action. \vspace{1mm}

\noindent \textbf{FF Planner}: In order to generate plans from the
metric version of the domain, we chose to use the metric version of
the Fast-Forward (FF) planner~\cite{hoffmann2003metric}. FF has been
used very successfully in the automated planning community for the
past decade, and can be run with a number of options that direct the
search process of the planner. The one that we found most pertinent to
our system was the `optimality' flag, which can be set (by default, it
is turned off). FF also keeps track of and outputs relevant statistics
when a plan is generated, like the number of search nodes that are
generated and expanded. This is helpful when measuring the amount of
planning effort required against problem instance and goal-set size.
\vspace{1 mm}

\noindent \textbf{LPG Planner}: We used the Local Plan Graph (LPG)
planner~\cite{gerevini2003planning} to generate temporal plans from
the version of the domain that was annotated with execution times on
the various actions. LPG is based on a local-search strategy, and can
return temporal plans (which may or may not be optimal with respect to
the duration of the resulting plan) in extremely short amounts of
time. This suited our application, since we needed a quick turnaround
on plan {\em generation}, and the plan quality itself isn't an
overriding consideration because of our plan-sense-replan
paradigm. Along with the generated plan for a given instance, LPG also
returns information such as the makespan (duration) of the plan, and
the start times of each action. \vspace{1 mm}

\noindent \textbf{Plan Container}: Once the respective planners
generate a plan that is appropriate for a given strategic planning
round, that plan needs to be parsed into a form that can be handed
over to the tactical planner and further on through the integrated
system. The important components that we parse out of a generated
plan, for each action, are: (1) the action name / label; (2) the
parameters, or objects, that instantiate that action instance; (3) if
the plan (and planner) is temporal, then the meta-information that
deals with the duration of that action, and the time that the action
is supposed to start executing. This information is parsed via a
specific parser that is unique to each planner. The information is
then stored in data structures so that it may be accessed by other
components as required.  The plan container is also tasked with
``remembering'' plans from previous rounds and determining the
differences (cancelations or new actions) to be taken as a result of
the current round. \vspace{-2mm}

\subsection{Tactical Planner}

The actions selected for execution by the plan container are then
passed to the tactical planner, which configures and runs analytics
corresponding to these actions. More specifically, for each action,
the tactical planner generates an analytic flow, and runs the flow to
completion. The analytic flows are deployed and run on analytic
platforms, e.g., IBM InfoSphere
Streams\footnote{\url{http://www-01.ibm.com/software/data/infosphere/streams/}}
and Apache Hadoop\footnote{\url{http://hadoop.apache.org/}}. As
implied by its name, the tactical planner uses its own planning to
configure the analytics and connect them into analytic flows. It first
translates each action received from the plan container into a
(tactical) planning goal, and generates a (tactical) plan, which it
then translates into an analytic flow. In our system, we separated the
tactical and strategic planners based on significant differences in
action semantics. During an investigation, multiple high-level sensing
actions may be executed, and the knowledge obtained by sensing may
affect the plan of the investigation. The tactical planner creates data
analysis plans for each action of the strategic plan. The resulting
tactical plans include lower-level instructions for configuring
analytics and data sources, and are never modified during the
execution process itself. \vspace{1 mm}


\noindent \textbf{MARIO}: Our implementation of the tactical planner
is similar to the MARIO~\cite{bouillet2009mario} system. In
particular, it accepts goals specified as a set of keywords (i.e.,
tags) and deployment parameters, and composes and deploys analytic
flows that meet these goals. Like MARIO, our tactical planner also
supports deployment of composed flows on a variety of platforms. In
our experiments, all analytics were either stream processing analytics
deployed on IBM InfoSphere Streams or on Apache Hadoop. We note that
for the purposes of the experiments described in this paper, it may
have been possible to build a system without a tactical planner --
replacing it with a simpler analytic launcher mechanism instead.
However, this would have required the manual wiring of each of the
analytic flows corresponding to a strategic action.
\vspace{1 mm}

\noindent \textbf{Simulator}: Our integrated system features a novel
tactical planning {\em simulator}, whose objective is to simulate
analytics for the automated planner and the larger strategic planning
problem. The idea behind designing and incorporating this simulator
module as part of the integrated system was to push the boundaries of
automated planning systems, and to generate progressively larger and
more complex problem instances. One of the main aims of our work was
to evaluate the best automated planners for different situations that
systems are likely to encounter in real world data, and the simulator
provides a way of quickly generating different kinds of real world
instances to enable this. In this module, we simulate the {\em sensing
  actions} that are at the core of the plan-sense-replan loop by
randomly generating effects from a pre-determined set of possible
effects (which can be obtained from domain experts). For example, a
hostset under investigation can be anomalous with respect to one of
three protocols; one of the simulator actions would thus be to assign
such a protocol at random (according to a pseudo-random generator
whose parameters can be set). The simulator then generates facts and
objects that are relevant to such an assignment, and inserts them back
into the world state. In this way, the simulator skips the actual
scheduling of decisions for execution on the analytics platforms, and
the subsequent gathering of data from the world, and provides a
quicker way of testing different automated planners on various
quasi-real instances. \vspace{-1.5mm}

\input{table-5-overview}

\subsection{Analytics Platforms}

Analytics that implement sensing actions are deployed in a scalable
computational infrastructure managed by distributed middleware.
Generally, we aim to support Big Data scenarios, where large clusters
of commodity computers can be used to process large volumes of stored
and streaming data. One such middleware platform for Big Data is IBM
InfoSphere Streams. For constructing historical models of behavior from
network data, we typically use the Apache Hadoop platform, by implementing
our analytics in Apache Pig.\footnote{\url{http://pig.apache.org}} \vspace{-1.5mm}

\subsection{Metadata Store}

The metadata store is a module that stores information about the raw
data coming in from the world, and the mappings from that raw data
into the representation that the automated planner uses (hence the
``meta'' in its name). When the analytics that are scheduled by the
tactical planner finish executing in the world, the results of these
computations are published to the metadata store. The mappings between
the various parameters are then utilized in order to interpret the
changes in the world into a planner-readable form. These are then
passed on to the state manager. \vspace{1.5mm}

\noindent \textbf{State Manager}: The state manager accepts changes in
the world that are passed on from the metadata store, and puts them
into a format that can be used as input to an automated planner. The
state manager is the last component in the plan-sense-replan loop: it
turns a deterministic, single iteration planner into the center of a
replanning system by creating a new problem instance based on the
execution of the plan generated in response to the previous
instance. The state manager also keeps track of the overall scenario
goals, and whether they have been achieved - in which case the
system's work is complete.
\vspace{-2mm}


\section{Evaluations}
\label{sec:5-evaluations}

\begin{figure*}[t!]
\begin{tabular}{MMM}
\footnotesize
\includegraphics[keepaspectratio, width=0.32\textwidth]{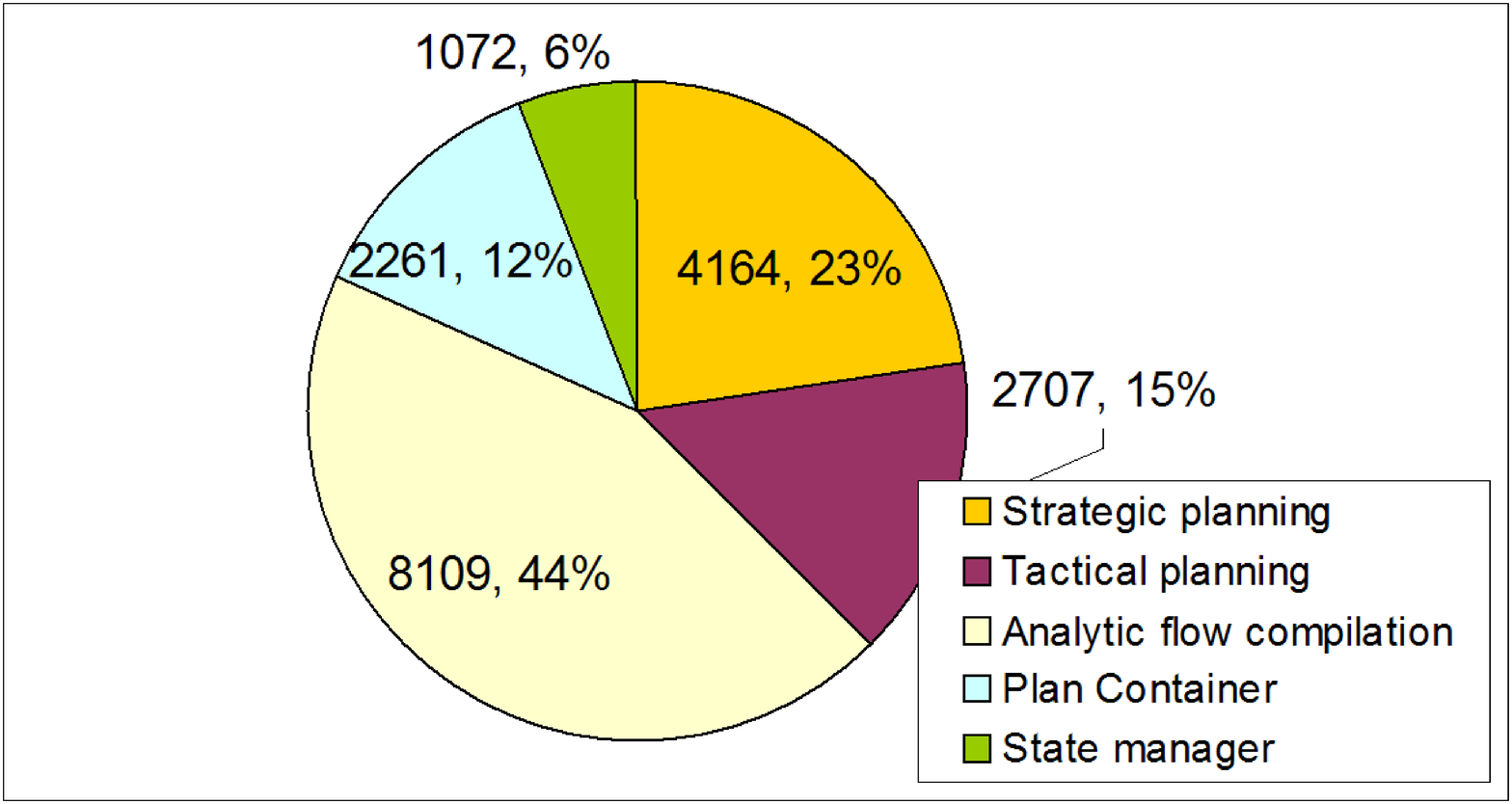} &
\includegraphics[keepaspectratio, width=0.32\textwidth]{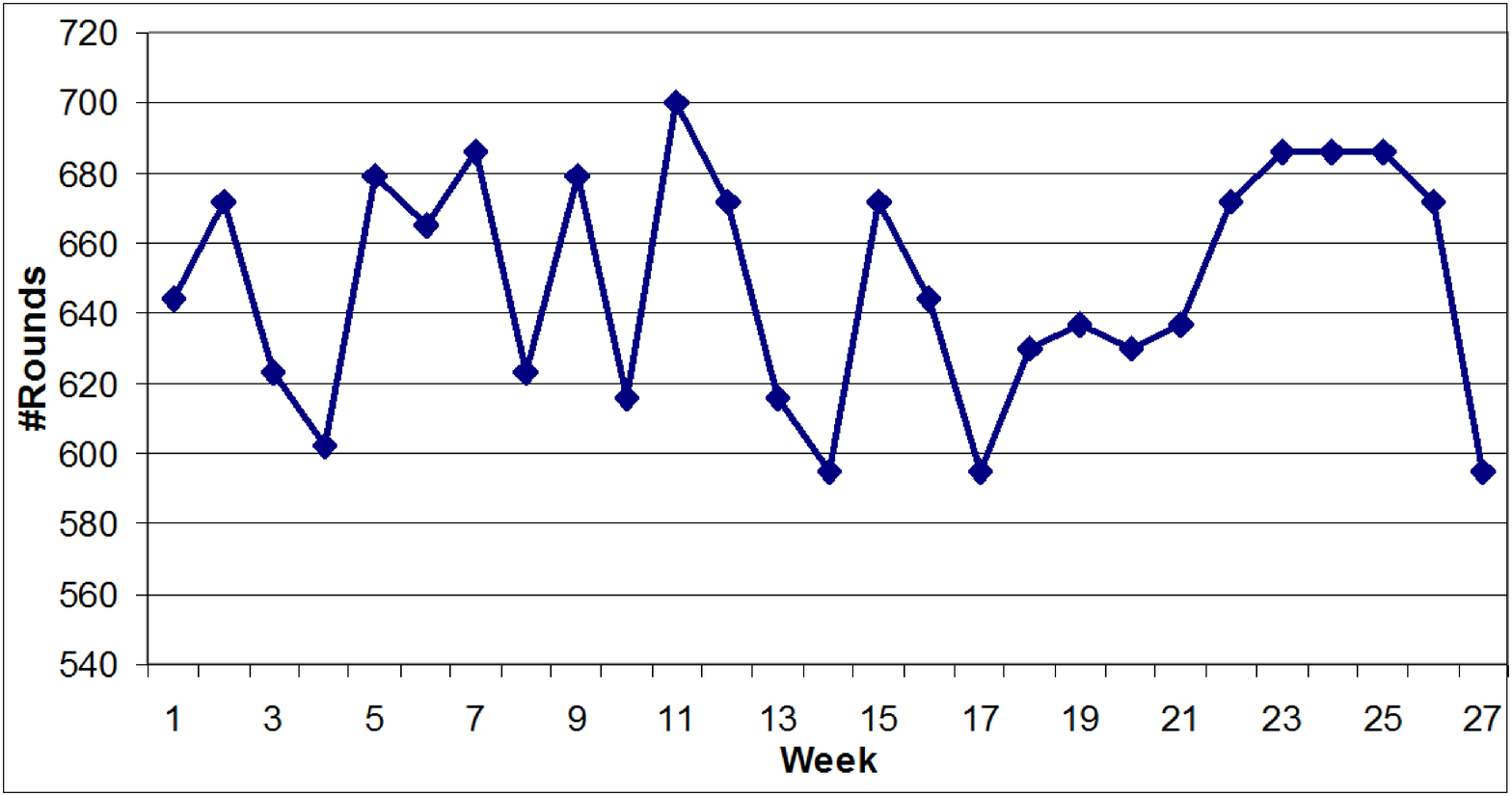} &
\includegraphics[keepaspectratio, width=0.32\textwidth]{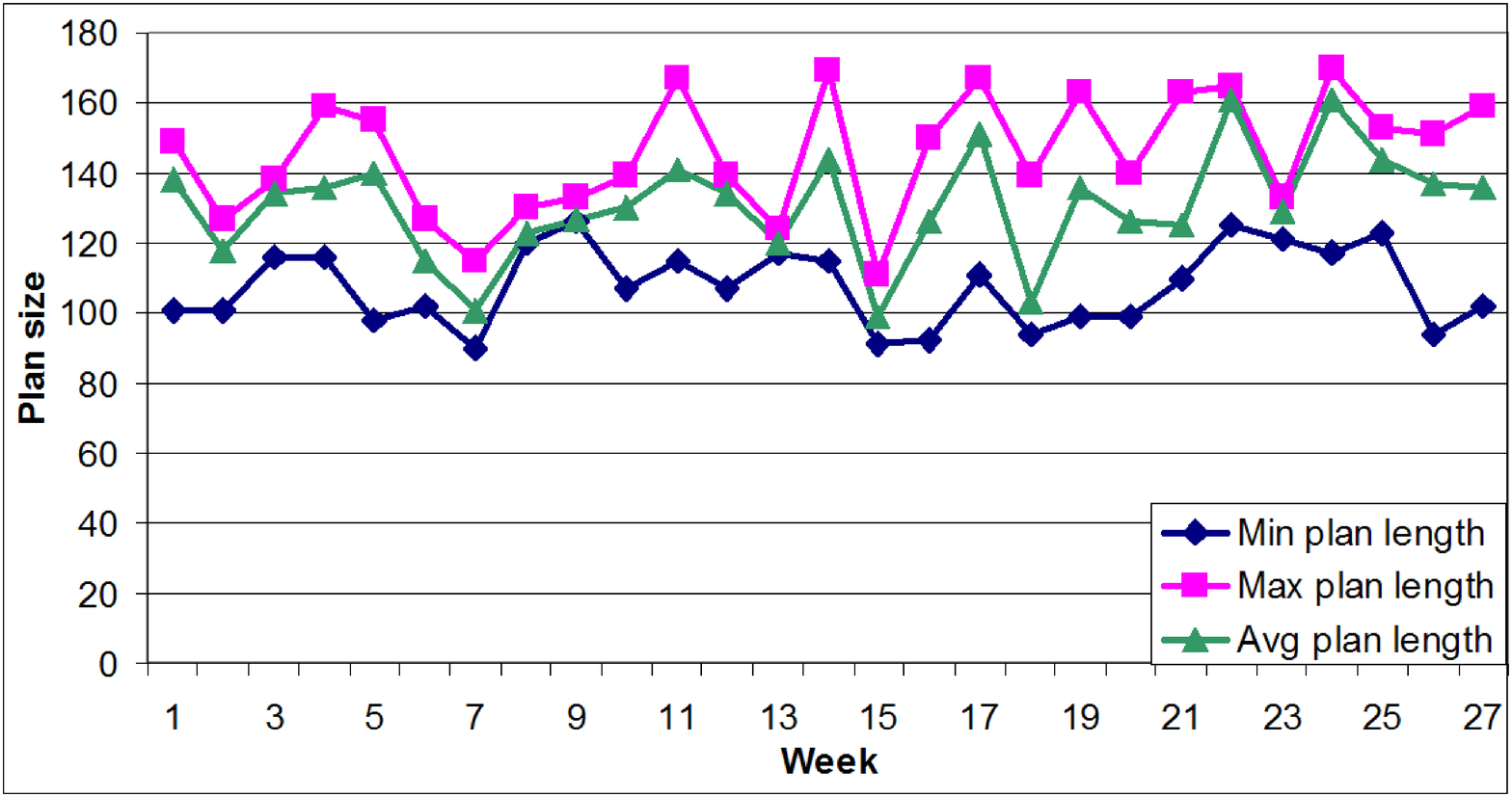} \\
(a) Timing (except analytic execution) & (b) Strategic planning rounds/week & (c) Strategic plan size/week \\
\includegraphics[keepaspectratio, width=0.32\textwidth]{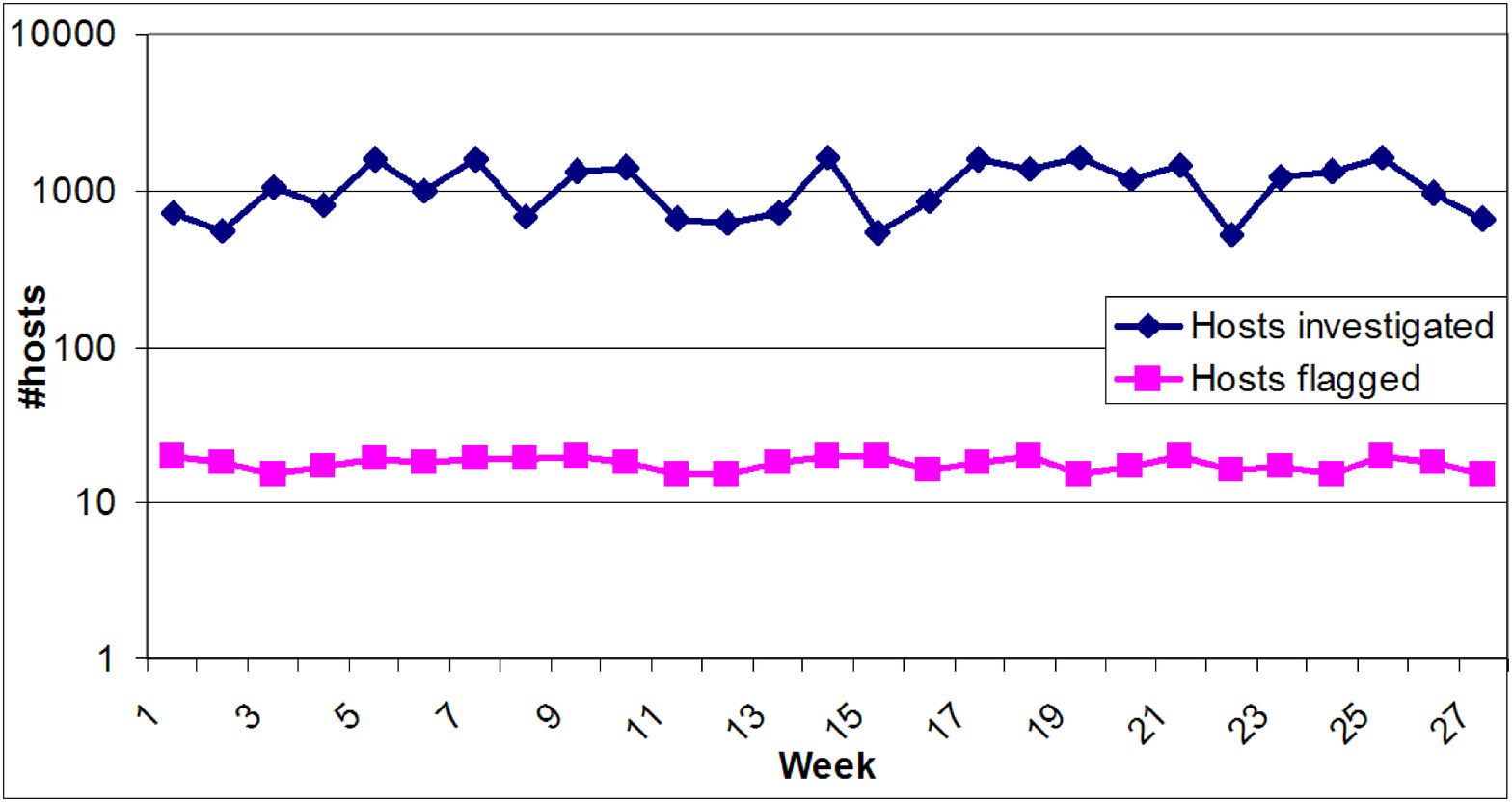} &
\includegraphics[keepaspectratio, width=0.32\textwidth]{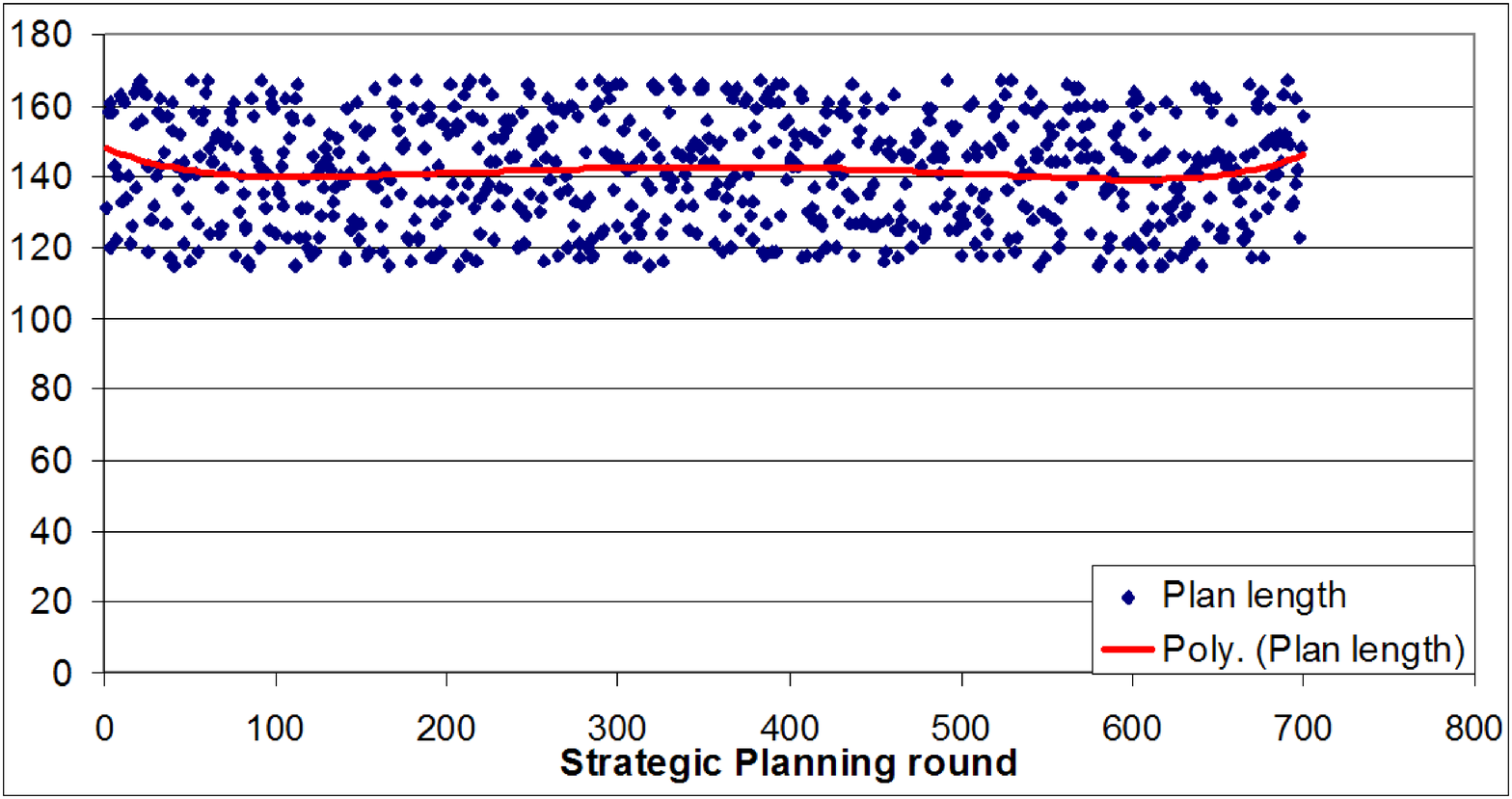} &
\includegraphics[keepaspectratio, width=0.32\textwidth]{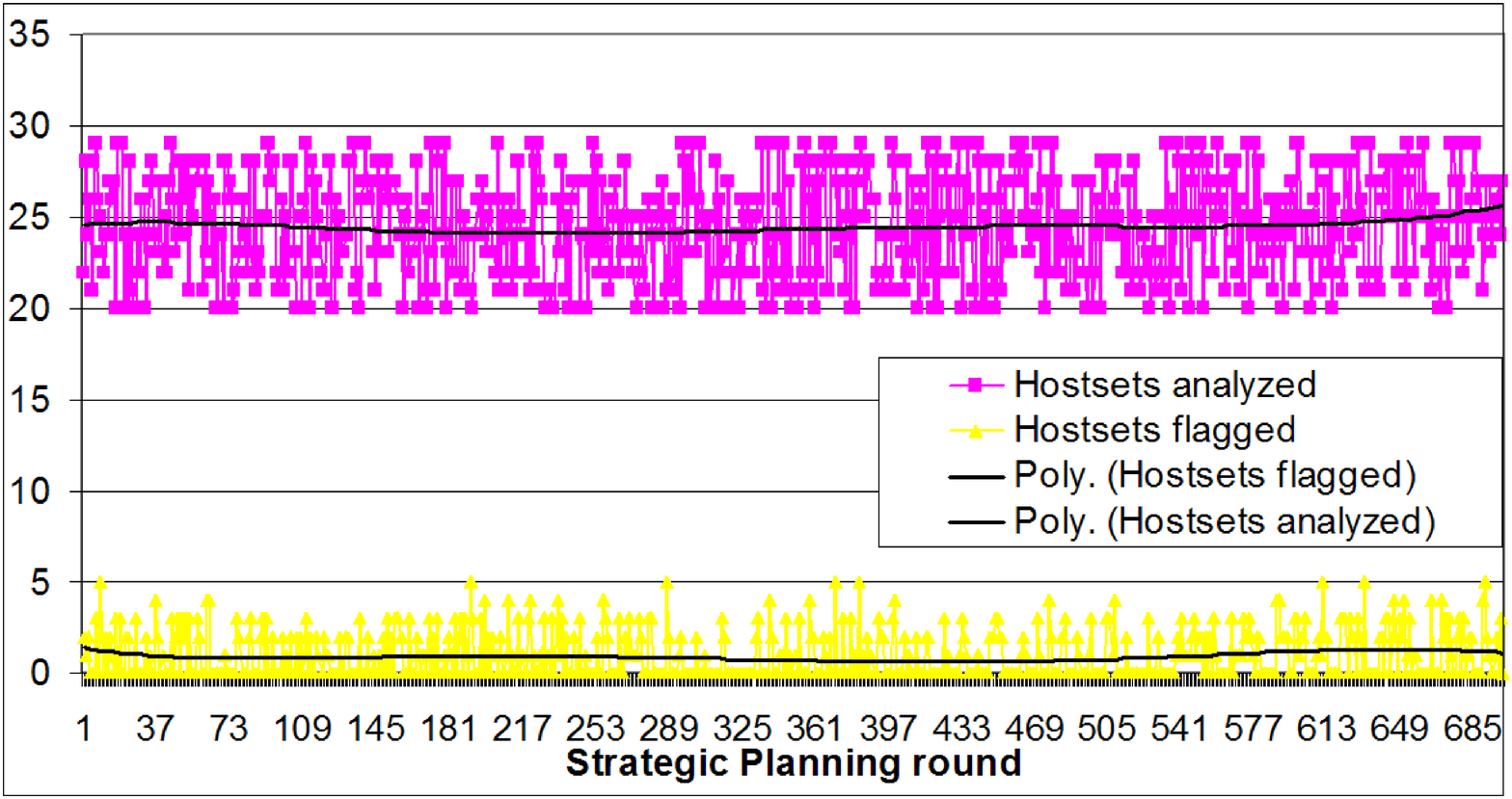} \\
(d) Hosts analyzed/flagged per week & (e) Plan sizes for one week (03/11-03/17) & (f) Hostsets analyzed/flagged for one week \\
\end{tabular}
\vspace{-0.02in}
\caption{Evaluation on network data}
\label{fig:res1}
\end{figure*}

We evaluated the integrated system and the domain we created for
network data analysis on recorded data from an institutional network,
covering a period of six months from January 1st 2012 through June
30th 2012. The data consisted of recorded traces for 19564 unique IP
addresses from the network for the following protocols:

\begin{itemize}
\item DNS requests and responses from all machines. \vspace{-0.0mm}
\item Netflow summaries of data going through the institutional
  firewall, sampled at 1\%. \vspace{-0.0mm}
\item IPFIX summaries of network traffic through the firewall, as well
  as internal to the network, sampled at 1\%. \vspace{-0.0mm}
\item sFlow (Sampled Flow) data, consisting of sample packets going
  through the firewall, sampled at 0.1\%. \vspace{-0.0mm}
\item Public information about blacklisted domains on the Internet
  from Google SafeBrowsing
  service\footnote{\url{https://developers.google.com/safe-browsing/}},
  as well as from the Malware Domain
  Blacklist\footnote{\url{http://www.malwaredomains.com/}}. \vspace{-0.0mm}
\end{itemize}

The total recorded data size, compressed, was approximately 342 GB. We
replayed this data using the IBM InfoSphere Streams middleware, at the
maximum possible replay speed. This resulted in a total execution time
of approximately 76 hrs and 23 minutes, which includes strategic and
tactical planning rounds, as well as the time taken for the planned
analytic flows to execute on InfoSphere Streams and Apache Hadoop. Our
network data analysis domain model is designed to perform a drill-down
analysis of host sets with anomalous behavior (e.g., substantially
higher amounts of traffic or traffic with a lot of geographically
distributed external locations in a short period of time). The
drill-down process is modeled in the planning domain to first identify
relatively large sets of hosts that exhibit some anomalous behavior at
a coarse level -- for instance, by looking at the total amount of
traffic -- and then to progressively refine these sets into smaller
and smaller subsets by performing more in-depth analysis -- for
instance, specific analyses based on particular protocols that are
identified as contributing to the anomalous behavior. As a result, our
strategic plans typically begin by considering all the hosts in the
network, identifying subsets that ``look'' anomalous, then narrowing
these down to subsets for which we can perform more specific analyses,
until finally we arrive at a very reduced set of hosts which we use to
inform a network administrator.

Each action in the strategic plan corresponds to a goal for a tactical
planner, which puts together the necessary analytic components to
achieve the analysis results for the corresponding step in the
strategic plan. Consequently, we implemented a tactical planning
domain where actions consist of analytic components which are put
together by the tactical planner to generate code for either
InfoSphere Streams or Apache Hadoop. Our tactical domain consists of
49 InfoSphere Streams components and 16 Apache Pig components. These
can be combined in 87 different programs (generated from the
corresponding plans), but for the purposes of our experiments, only 33
of these had strategic actions associated with them (but note that
since these can have runtime parameters, the number of possibilities
is actually much greater). During our evaluation, the tactical planner
deployed approximately 31,000 analytic flows. \vspace{1 mm}

\noindent \textbf{Timing}: We measured the total time spent for
strategic planning, tactical planning (including compilation of
generated code) and other integrated system components. The results
are shown in Figure \ref{fig:res1}(a). The important thing to notice
here is that the combined overhead of the integrated system accounts
for 6.65\% of the total execution time; hence it is
negligible.\vspace{1 mm}

\begin{figure}[h!]
\centering
\includegraphics[keepaspectratio,
width=0.45\textwidth]{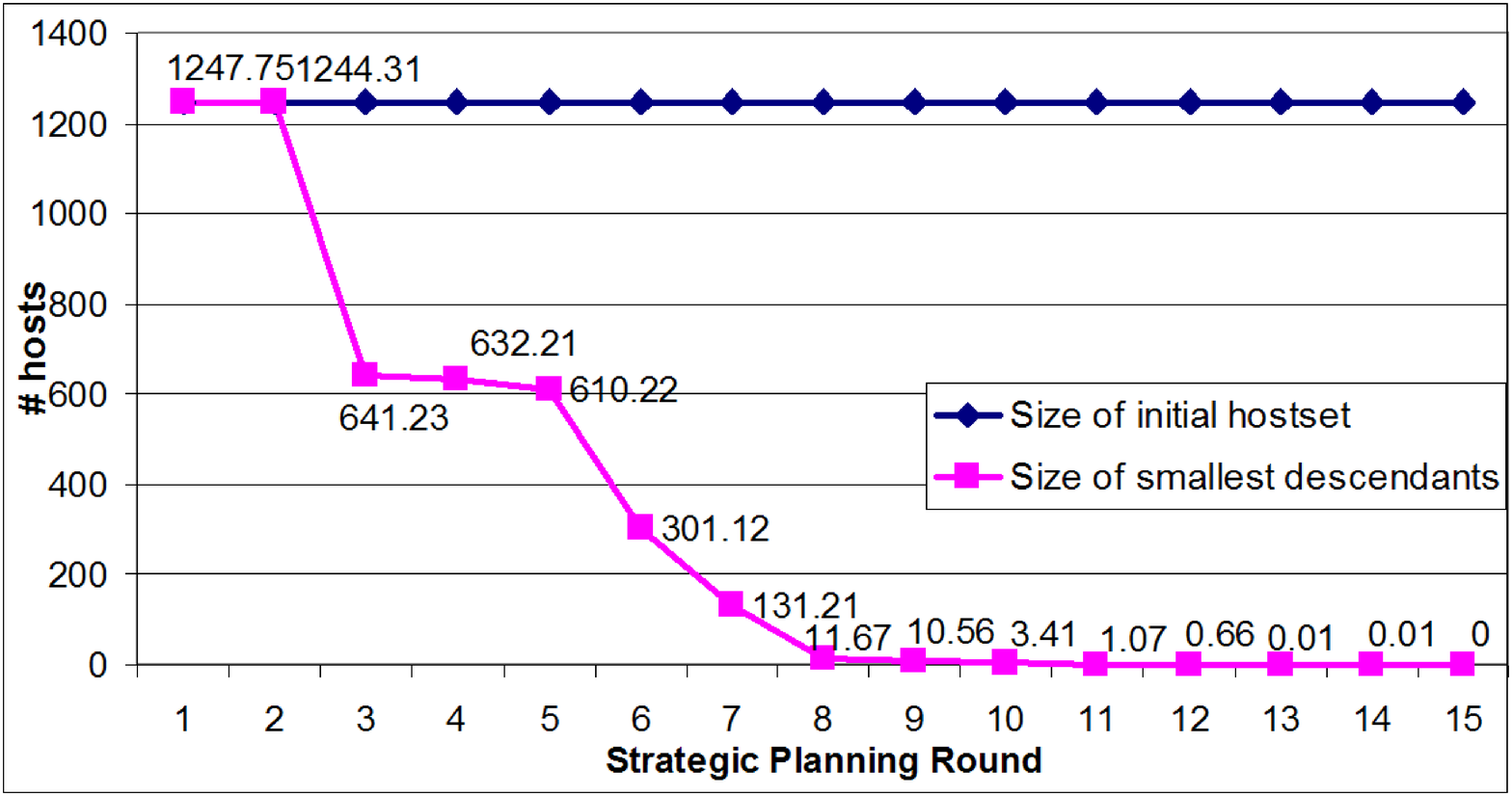}
\vspace{-0.05in}
\caption{Drill-down effectiveness}
\label{fig:res2}
\end{figure}

\noindent \textbf{Strategic planning}: We looked at a few measures
related to the strategic planning process. First, we counted how many
changes to the state of the world (and hence, reruns of the strategic
planner) are typically required. From Figure \ref{fig:res1}(b), we can
see that the number is relatively stable around 640 rounds per week of
data. Some ups and downs are due to seasonal effects of network
activity (e.g., day/night, week/weekend cycles), as well as expiration
and rebuilding of network traffic models based on which we identify
anomalous hosts. The same stability is evident in the plan sizes,
which lie in a tight band around 120 actions (Figure
\ref{fig:res1}(c)). We should point out that since we replan for every
change in the state of the world, some of the actions during the
replanning may already be running (their corresponding analytic flows
are deployed); others may be speculative paths that require the
execution of a sensing action to decide whether that path will indeed
be taken or replanning is necessary. The true number of changes
(canceling or deploying analytic flows for stopped or new strategic
actions) on every replan averages to just 6.1. Finally, we have chosen
a week at random (in this case, week 11) and plotted the plan sizes in
Figure \ref{fig:res1}(e). This indicates some local jitter, but the
trend is that of a stable plan size. We noticed that at this scale,
there is no noticeable effect of typical activity cycles (day/night
for instance). We believe this is because of two reasons: (i) inertia
in analyzing each host set -- once such a set is identified, it takes
a few replanning rounds for this set to be refined to something that
can be sent to an administrator; and (ii) models of network traffic
are periodically rebuilt, creating larger plan sizes when this
occurs. In this model, most models expire after 24 data
hours.\vspace{1mm}


\noindent \textbf{Hosts analyzed}: The number of hosts and host sets
analyzed follows a similar stable trend. This is primarily due to the
way the domain is modeled as a drill-down process in which each
hostset will eventually be either sent to an admin in its most refined
form or will become empty and hence be discarded. We should point out
that modeling the domain using host sets may cause an exponential
blowout in the number of objects in the planning domain. We see a
limited form of this behavior in this domain, where we analyze a little
over 19,000 hosts, but over 390,000 host sets during the entire experiment. Figure
\ref{fig:res1}(d) shows the number of unique hosts investigated (that
were in any host set in the planning domain during that data week) and
the number of unique hosts flagged in every week; uniqueness here
refers to the respective week only -- although we send 371 unique
hosts to the administrator, each host is on average flagged 4.96
times. Figure \ref{fig:res1}(f) shows a one week detail for the number
of hostsets analyzed and flagged within a week.\vspace{1mm}

\noindent \textbf{Effectiveness of the planning approach}: Our
approach to analyzing network data is primarily aimed at making the
task of the network administrators easier. In that respect, we looked
at the set of hosts that are ultimately flagged as abnormal by our
system and compared them to what was flagged by a commercial security
appliance in the same network in the same interval (a Cisco PIX 500
series device). We found that 187 out of the 371 hosts flagged by our
system were also flagged by the security device in the same
interval\footnote{As an aside, security appliances usually flag hosts
  that display violation of a predetermined set of rules that refer to
  a small number of packets, and not abnormal behavior over larger
  periods.}.  The security device also flagged an additional 127 hosts
that our system did not catch; of the remaining hosts that our
planning approach flagged, manual inspection revealed more than a
dozen which manifest Web crawling behavior, and further analysis is
needed for the rest. This leads us to conclude that the automated
planning approach is in the comparable range of typical security
appliances, and it simplifies the network monitoring task by
intelligently guiding the application of complex analytics on a
smaller set of data than commercial security appliances (extrapolating
from sampled packets, the security appliance discussed analyzed two to three 
orders of magnitude more data). Furthermore, we believe
refinement of the planning domain we created (beyond the three
protocols that were analyzed in depth) would further improve results.

We also analyzed the way the planning domain developed is able to
perform an effective drill-down analysis of hosts. Over a number of
rounds, we considered the size of the first host set (anomalous at a
coarse level) versus the total size of the most refined subsets
obtained up to that point. The results in Figure \ref{fig:res2} show
that typically a host set is completely analyzed within 14 -- 15
rounds of strategic planning.

\vspace{1 mm}
\noindent \textbf{Model Scalability}: Since one of the contributions
of our work was the novel PDDL model of the network administration
scenario that we developed, we decided to evalute the scalability of
that model on different planning systems.  Such experiments also
allowed us to observe the domain's behavior independent of the other
integrated system components (the experiments described
previously). In order to produce these results, we generated problem
instances of increasingly large sizes -- where size is either one of
(1) the number of initial hostsets, or (2) the number of goals in the
problem instance. We then provided these problem instances, along with
the domain model, to both the FF (metric version) and LPG (temporal
version) planners. Additionally, we also ran a subset of the problem
instances with the SPPL planner\cite{riabov2006scalable}.

\section{Conclusions and Future Work}

In this paper, we presented the novel strategic planning problem for
the automation of real world safety-critical applications and
described the creation of a new PDDL model that captured one such
domain related to network administration and monitoring. We then
discussed the application of an integrated system centered around an
automated planner's decisions to this model and real problems from
that network administration domain, and presented promising results on
a real world network.

As demonstrated previously, our integrated system does a reasonable
job of automating the strategic planning process. However, much
remains to be done in terms of testing the full capabilities of the
system. On the modeling side, we are currently looking at modeling
other business applications that feature large amounts of streaming
data as well as sensing actions. One such application that we have
started modeling is data analysis for healthcare in an intensive care
setting, where data is gathered from the various instruments that are
deployed on a patient, and investigations must be launched based on
existing medical theories as well as the histories of past patients.
We are also looking to expand the capabilities of the integrated
system in at least three major ways: (1) incorporate more automated
planners into the system, so that we may choose the best one depending
on the particular scenario and problem instance at hand; (2) create a
comprehensive report on the performance of different automated
planners under different problem instances; and (3) tuning the
parameters of the underlying analytic processes automatically during
execution.


\bibliographystyle{aaai}
\bibliography{master}

\end{document}

%% file: table-5-overview.tex
\begin{table}[t!]%
\footnotesize
\begin{tabular}{|l|p{.3\textwidth}|}
\hline
\textbf{Data interval} & 1/1/2012 -- 1/7/2012 \\\hline
\textbf{Data used} & DNS, Netflow, IPFIX, sFlow, external blacklists \\\hline
\textbf{Data size} & 342 GB compressed \\\hline
\textbf{Execution time} & 76h 23m 12s \\\hline
\textbf{Tactical planner} & \parbox{0.3\textwidth}{49 Streams, 16 Apache Pig components;\\ approx. 31000 analytic flow deployments} \\\hline
\textbf{Strategic planner} & 17280 rounds (plans) \\\hline
\textbf{Hosts} & 371/19564 unique IPs flagged, 390000 hosts set analyzed \\\hline
\end{tabular}
\caption{Evaluation summary}
\label{tab:eval-summary}
\end{table}
\newcolumntype{M}{>{\centering\arraybackslash}m{\dimexpr.32\linewidth-2\tabcolsep}}